\documentclass[11pt]{article}

\usepackage[final]{acl}

\usepackage{times}
\usepackage{latexsym}
\usepackage{listings}
\usepackage{amsmath}

\usepackage[T1]{fontenc}

\usepackage[utf8]{inputenc}

\usepackage{microtype}

\usepackage{inconsolata}

\usepackage{graphicx}

\usepackage{booktabs}

\newcommand{\customfootnotetext}[2]{{
  \renewcommand{\thefootnote}{#1}
  \footnotetext[0]{#2}}}

\title{More Human, More Efficient: Aligning Annotations with Quantized SLMs}

\author{Jiayu Wang\textsuperscript{\textbf{*}} \and Junyoung Lee\textsuperscript{\textbf{*},\textbf{†}} \\
  Home Team Science and Technology Agency\\
  \texttt{\{lastname\}\_\{firstname\}@htx.gov.sg} 
  }

\begin{document}
\maketitle
\begin{abstract}

As Large Language Model (LLM) capabilities advance, the demand for high-quality annotation of exponentially increasing text corpora has outpaced human capacity, leading to the widespread adoption of LLMs in automatic evaluation and annotation. However, proprietary LLMs often exhibit systematic biases that diverge from human expert consensus, lacks reproducibility, and raises data privacy concerns. Our work examines the viability of finetuning a quantized Small Language Model of 1.7B parameter size on limited human-annotated data to serve as a highly aligned, deterministic evaluator and annotator. By implementing a custom, multi-dimensional rubric framework and simple augmentation and regularization techniques, the proposed approach achieves higher inter-annotator agreement (0.23 points increase in Krippendorff’s $\alpha$) than the best performing state-of-the-art proprietary LLM. We also demonstrate the generalizability of the proposed training pipeline on a separate emotion classification task. The results show that task-specific alignment and efficient 4-bit quantized fine-tuning provide superior open-source alternative to using proprietary models for evaluation and annotation. Our finetuning approach is publicly available at \url{https://github.com/jylee-k/slm-judge}.
\end{abstract}

\customfootnotetext{\textbf{*}}{Equal contribution}
\customfootnotetext{\textbf{†}}{Corresponding author}

\section{Introduction}

The recent proliferation of Large Language Models (LLMs) has shifted the focus of natural language processing from discriminative classification to complex generative tasks, requiring rigorous and scalable evaluation frameworks. Traditional reference-based lexical metrics, such as BLEU and ROUGE, are increasingly recognized as inadequate for assessing modern requirements like semantic nuance, stylistic alignment, and factual consistency of candidate LLMs. Research focus has shifted toward automatic evaluation in an LLM-as-a-Judge (LaaJ)\footnote{Although the term LaaJ typically refers to LLMs evaluating outputs of other LLMs, it can be viewed as a special case of the broader “LLM-as-an-Annotator” paradigm. Since LaaJ is more widely used, it is used as the consistent terminology throughout this work to refer more generally to any evaluation, annotation, or labeling traditionally performed by humans~\cite{calderon-etal-2025-alternative}.} setting, where powerful proprietary LLMs are prompted to evaluate the outputs of candidate systems based on zero-shot or few-shot rubrics~\cite{li-etal-2025-generation}. A similar trend of using LLM-as-an-Annotator is observed~\cite{tan-etal-2024-large}, leveraging on the scalability and speed of LLMs on annotating candidate text.

However, reliance on proprietary models introduces significant vulnerabilities. Despite their sophisticated reasoning capabilities, commercial APIs are black-box systems characterized by opaque versioning and API deprecations which undermines reproducibility, and concerns of data sovereignty in sensitive domains like legal or medical research. Moreover, these models harbor deeply ingrained evaluation biases such as position bias~\cite{wang-etal-2024-large-language-models-fair, shi-etal-2025-judging}, verbosity bias~\citep{huang-etal-2025-empirical, park-etal-2024-offsetbias, ye2025justice}, bias for response with lower perplexity~\citep{wataoka2024self}, or preference for a more visually appealing response~\citep{chen-etal-2024-humans}. The performance of LaaJ also varies significantly across tasks~\citep{wang2025can}. In the context of linguistic annotation, these machine-driven biases and performance deviations manifest as objective annotation errors that distort the true performance of the target and misinform subsequent model training.

These systemic flaws undermine the reproducibility and reliability of automated evaluation and annotation, creating a need for transparent, open-source alternatives. In this work, we address these shortfalls by demonstrating that a supervised finetuning pipeline---utilizing a quantized Small Language Model (SLM) and targeted data augmentation on limited, high quality human annotations---can provide better aligned and reproducible automatic annotations compared to proprietary models. 

\section{Experimental Setup}\label{sec:expt-setup}

\subsection{Rubric Development}\label{ssec:rubric-development}

To examine the LaaJ capabilities fairly, we wish to reduce the possibility of the proprietary models having been exposed to certain metrics during training time. Hence, we adopt the three high-level criteria set out for evaluating natural language generation in RankME~\cite{novikova-etal-2018-rankme}---naturalness, quality, and informativeness---decompose them into a more granular approach where explicit evaluation criteria were defined for each metric, inspired by \citet{amidei-etal-2019-use} and \citet{biyani2024rubicon}. 

\begin{figure}[h]
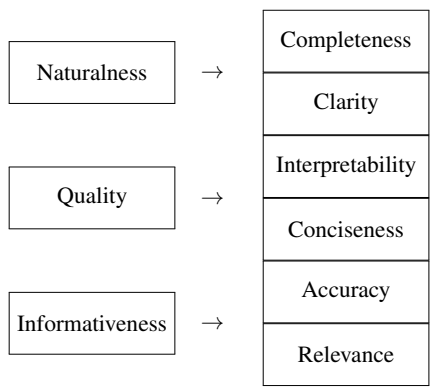

\centering
\scalebox{0.8}{
\renewcommand{\arraystretch}{1.3}
\setlength{\tabcolsep}{6pt}
    \begin{tabular}{c c c}
    \fbox{\parbox[c][0.8cm][c]{2.5cm}{\centering Naturalness}} & $\rightarrow$ & 
        \begin{tabular}{c}
        \fbox{\parbox[c][0.8cm][c]{2.5cm}{\centering Completeness}} \\[0.2cm]
        \fbox{\parbox[c][0.8cm][c]{2.5cm}{\centering Clarity}}
        \end{tabular} \\[0.5cm]
    
    \fbox{\parbox[c][0.8cm][c]{2.5cm}{\centering Quality}} & $\rightarrow$ & 
        \begin{tabular}{c}
        \fbox{\parbox[c][0.8cm][c]{2.5cm}{\centering Interpretability}} \\[0.2cm]
        \fbox{\parbox[c][0.8cm][c]{2.5cm}{\centering Conciseness}}
        \end{tabular} \\[0.5cm]
    
    \fbox{\parbox[c][0.8cm][c]{2.5cm}{\centering Informativeness}} & $\rightarrow$ & 
        \begin{tabular}{c}
        \fbox{\parbox[c][0.8cm][c]{2.5cm}{\centering Accuracy}} \\[0.2cm]
        \fbox{\parbox[c][0.8cm][c]{2.5cm}{\centering Relevance}}
        \end{tabular}
    \end{tabular}
}
\caption{Hierarchical structure of the proposed evaluation rubric.}
\label{fig:rubric_structure}
\end{figure}


Each dimension is scored on an ordinal scale from $\{-2, -1, 0, 1, 2\}$, where $-2$ indicates a severe failure to meet the criterion, and $2$ indicates perfect satisfaction of the metric. A complete definition of the rubric boundaries can be found in Appendix
~\ref{subsec: rubric description}.

\subsection{Dataset Curation}

In examining domain specificity, we curate a specialized question-and-answer (QnA) dataset based on web-scraped data from the Singapore Prison Service (SPS) website\footnote{\url{https://www.sps.gov.sg/}}. The raw text spans multiple SPS website topics, including careers, rehabilitation programs, and annual reports, with contexts of varying lengths. The dataset consists of 97 human-written questions, each related to a text chunk from the website. For each question, we then generate 7 candidate responses from 7 different open-source SLMs (listed in Appendix~\ref{apdx:slms}) for variability of response quality.

Afterwards, expert human annotators manually score these responses across the 6 criteria, with 2 sets of scores per candidate response. While it is common to assume there is a ground truth label by having the annotators agree on a single label, we choose to preserve label diversity to preserve disagreements in human perception~\cite{weerasooriya-etal-2023-disagreement}. Since the evaluation relies on a multi-rater, ordinal scale where boundary ambiguity is present, we utilize Krippendorff's Alpha ($\alpha$), accounting for varying magnitudes of disagreement and multiple raters~\citep{krippendorff2013content}.

\begin{table*}[!h]
    \centering
    \begin{tabular}{ccccc}
        \toprule
          & Krippendorff’s $\alpha$  \\
        \midrule
        Our Method  & \textbf{0.5774} \\
        
        Full finetuning without augmentation  & 0.4304  \\
        
        Early stopping without augmentation  & 0.4380 \\
        
        LoRA dropout ($p = 0.1$)  & 0.4067 \\
        \midrule
        Zero-shot GPT-4o  & 0.1964  \\
        
        Zero-shot GPT-5-mini-2025-08-07  & 0.2462 \\
        
        Zero-shot GPT-5-nano  & 0.1065\\
        
        Zero-shot GPT-5.2-chat  & 0.2054  \\
        \midrule
        Few-shot GPT-4o  & 0.0101  \\
        
        Few-shot GPT-5.2-chat  & 0.0471  \\
        \bottomrule
    \end{tabular}
    \caption{Annotator agreement results on SPS dataset.}
    \label{tab: sps results}
\end{table*}

\subsection{Data Augmentation and Regularization}

Given the limited size of the human-annotated dataset, SLMs are highly susceptible to overfitting to the specific semantic phrasing of the training prompts~\cite{wang2024comprehensivesurveysmalllanguage}. To improve model generalization, we propose three specific data augmentation and regularization strategies.


First, we apply prompt paraphrasing to introduce syntactic variation. For example, changing the instruction from "You \textit{will} be given a context..." to "You \textit{are} given a context..." ensures the model learns the underlying evaluation logic rather than memorizing a rigid prompt template.


Second, we employ component permutation (similar to swap augmentation). We randomly permute the order of the input components (e.g., swapping the sequence of QUESTION, CONTEXT, and ANSWER) to combat inherent position bias, forcing the model to evaluate the text holistically.

Lastly, we implement token dropout. We randomly mask out non-essential tokens within the prompt during the training phase with a fixed probability. Token dropout acts as a powerful structural regularization technique~\cite{gao2025enhancingelusivecluesknowledge}, structurally similar to methods such as standard input noise injection, preventing the model from over-relying on specific lexical artifacts in the small training corpus.

An example of the input prompt can be found in Appendix~\ref{apdx:sps-laaj-prompt}. Train-test split of 90-10 is used, after the dataset is augmented with above mentioned techniques.

\subsection{Training Pipeline}

The proposed training pipeline uses the Unsloth~\citep{unsloth} library to perform 4-bit quantized parameter-efficient fine-tuning on an SLM --- Qwen3-1.7B which is lightweight and can be hosted on a single consumer-grade GPU or an edge device. The training hyperparameters can be found in Appendix~\ref{apdx:hyperparams}.

We formulate the evaluation as a causal language modeling problem rather than a ordinal classification task, by appending the 6 scores from human annotation directly to the prompt as a completion string, such that the model is trained to generate the annotations as a logical extension of its reasoning over the candidate response. We employ a completion-only loss\footnote{\url{https://huggingface.co/docs/trl/sft_trainer\#train-on-completion-only}} during training, such that the model is only trained to predict the scores based on the given rubric.

\subsection{Baselines}

We benchmark our model against the following:
\begin{itemize}
    \item Zero-shot prompting: GPT-4o, GPT-5-nano, GPT-5-mini-2025-08-07, and GPT-5.2-chat
    \item Few-shot prompting of GPT-4o and GPT-5.2-chat---the prompt is finetuned via MIPROv2 optimizer~\citep{opsahl-ong-etal-2024-optimizing} using \texttt{dspy}~\citep{khattab2023dspycompilingdeclarativelanguage}
    \item Variants of proposed PEFT pipeline: without proposed augmentation; with LoRA dropout
    
\end{itemize}

The base model for Qwen3-1.7B could not generate target labels even after extensive finetuning, and hence is not selected as baseline. Finetuned models, including our proposed approach, are only run once because of their deterministic nature. For non-deterministic models (i.e. commercial APIs), we conduct 3 independent annotation runs for each model and report the average score.

\begin{table*}[!htbp]
    \centering
    \begin{tabular}{ccccc}
        \toprule
         & Accuracy & Macro-F1 \\
        \midrule
        Our Method & $0.8163$ & $0.6380$ \\
        
        Full SFT without Augmentation & $0.7819$ & $0.4967$  \\
        
        Early Stopping without Augmentation & $0.7917$ & $0.5685$ \\
        
        LoRA Dropout ($p = 0.1$) & $0.7937$ & $0.4998$ \\
        \midrule
        Zero-shot GPT-4o & $0.4741$ & $0.3732$ \\
        
        Zero-shot GPT-5-mini-2025-08-07 & $0.4732$ & $0.3875$ \\
        
        Zero-shot GPT-5-nano & $0.5177$ & $0.3969$ \\
        
        Zero-shot GPT-5.2-chat & $0.5062$ & $0.4099$ \\
        \midrule
        Few-shot GPT-4o & $0.4597$ & $0.2309$  \\
        \
        Few-shot GPT-5.2-chat & $0.4990$ & $0.3926$  \\
        \bottomrule
    \end{tabular}
    \caption{Classification performance of proposed approach and baselines on the GoEmotions dataset. Classification performance is reported in place of inter-annotator agreement as the dataset labels are taken as ground truth.}
    \label{tab: goemotions results}
\end{table*}

\section{Results}

The results on the SPS Dataset are shown in Table~\ref{tab: sps results}. Our proposed method achieves an $\alpha$ of 0.5774, representing a substantial improvement over all proprietary baselines. Specifically, our 1.7B SLM outperformed the best-performing proprietary model, GPT-5-mini-2025-08-07 ($\alpha=0.2462$), by 0.3312 points.

Despite their larger parameter counts, proprietary models like GPT-4o ($\alpha=0.1964$) and even \texttt{dspy}-optimized versions ($\alpha=0.0101$) failed to achieve moderate agreement with human annotators.

Comparisons against baselines without data augmentation (best $\alpha=0.4380$) demonstrate that our proposed augmentation and regularization strategies provided a significant increase in inter-annotator agreement. The training plots for training without augmentation and LoRA dropout~\citep{lin2024lora} can be found in Appendices~\ref{subsec: train/test loss plot augmentation} and \ref{subsec: train/test loss plot lora dropout}, further substantiating their effectiveness.




\section{Generalizability to Other Annotation Tasks}

We also evaluated our method on the GoEmotions dataset~\citep{demszky-etal-2020-goemotions}, keeping the same training hyperparameters, to show the generalizability of the proposed approach to an emotion classification task on text.

The results in Table~\ref{tab: goemotions results} confirm that the proposed approach is task-agnostic. Our method achieved an accuracy of 0.8163 and a Macro-F1 of 0.6380, nearly doubling the accuracy of GPT-4o (0.4741) and GPT-5.2-chat (0.5062). This suggests that for classification and labeling tasks, task-specific alignment on a small, high-quality dataset is more effective than the broad general-purpose reasoning of proprietary LLMs.

\section{Discussion}



The experimental results demonstrate that task-specific alignment using a quantized 1.7B parameter SLM consistently outperforms significantly larger proprietary models in both specialized linguistic evaluation and general emotion classification. While one might expect massive LLMs to excel due to their extensive pre-training, their reliance on zero-shot or few-shot prompting often fails to overcome inherent evaluation biases—such as position or verbosity bias—and lacks the precision required for niche, multi-dimensional rubrics. Our findings suggest that for abstract annotation tasks, a smaller model focused on a high-quality, task-specific dataset is more effective than a generalist "black-box" model designed for concrete reasoning.

\section{Conclusion}



Our work demonstrates that a quantized SLM evaluator, fine-tuned on limited human annotations using granular framework, can serve as a highly aligned and scalable alternative to proprietary models. Our results show that a 1.7B model can achieve significantly higher inter-annotator agreement than state-of-the-art proprietary LLMs, in both simple text classication task and specialized domains. The transition from black-box commercial APIs to locally hosted, fine-tuned SLMs resolves the core challenges of cost, data privacy, evaluation bias, and reproducibility, paving the way for more democratized and reliable AI-driven annotation.

\section*{Limitations}

We did not benchmark against proprietary models outside of the GPT series primarily due to lack of API access. From the list of models that we had access to, we had hoped to cover models across a range of parameter sizes and release periods by choosing GPT-4o, GPT-5-mini, GPT-5-nano, and GPT-5.2-chat.

\section*{Acknowledgments}

We would like to thank the project members from HTX and SUTD---Jane, Shisheng, Jason, Valerie, Wenxuan, Chen Huang, and James---for the valuable discussions over the course of the work.

\bibliography{custom}

\onecolumn
\appendix
\section{Appendix}
\label{sec:appendix}

\subsection{LLM Response Evaluation Rubric Description} \label{subsec: rubric description}

\begin{itemize}
    \item Completeness: \\
    -2: "Important information is missing, causing major misunderstandings." \\
    -1: "Several details are missing, making the response only partially usable." \\
    0: "Mostly complete but lacking a few supporting details." \\
    1: "Complete with all necessary information and minimal omissions." \\
    2: "Fully comprehensive with all required details and no omissions."
    \item Clarity: \\
    -2: "Very unclear and confusing, making it hard to understand." \\
    -1: "Partially unclear with awkward wording or ambiguous sentences." \\
    0: "Somewhat clear but with minor ambiguity or weak phrasing." \\
    1: "Clear, easy to follow, and well-phrased." \\
    2: "Extremely clear, well-articulated, and highly readable."
    \item Interpretability: \\
    -2: "Difficult to understand with tangled reasoning or unclear logic." \\
    -1: "Partially understandable but with unclear logic or weak organization." \\
    0: "Generally understandable but occasionally confusing or inconsistent." \\
    1: "Easy to understand with clear logic and strong organization." \\
    2: "Extremely easy to understand, logically strong, and excellently organized."
    \item Conciseness: \\
    -2: "Very wordy, redundant, or filled with unnecessary details." \\
    -1: "Somewhat verbose with noticeable redundancy." \\
    0: "Some unnecessary wording but overall acceptable length." \\
    1: "Concise with minimal redundancy and clear expression." \\
    2: "Highly concise, focused, and free of all unnecessary words."
    \item Accuracy: \\
    -2: "Contains factually incorrect or fabricated information." \\
    -1: "Contains several factual inaccuracies or unclear claims." \\
    0: "Mostly accurate but with minor errors or ambiguous statements." \\
    1: "Accurate and reliable with no significant factual issues." \\
    2: "Fully precise, factually correct, and verifiable throughout."
    \item Relevance: \\
    -2: "Content is mostly irrelevant or off-topic." \\
    -1: "Content is partially irrelevant or only loosely connected to the topic." \\
    0: "Content is somewhat relevant but contains unnecessary or unfocused parts." \\
    1: "Content is relevant and contributes meaningfully to the topic." \\
    2: "Content is highly relevant, targeted, and fully aligned with the topic."
\end{itemize}

\subsection{SLMs used for SPS dataset curation} \label{apdx:slms}
\begin{itemize}
    \item LGAI-EXAONE/EXAONE-4.0-1.2B
    \item meta-llama/Llama-3.2-1B-Instruct
    \item ibm-granite/granite-3.3-2b-instruct
    \item mistralai/Ministral-3-3B-Instruct-2512
    \item meta-llama/Llama-3.2-3B-Instruct
    \item Qwen/Qwen3-4B-Instruct-2507
    \item google/gemma-3-4b-it
\end{itemize}

\subsection{Example of Prompt for Finetuning on SPS Dataset} \label{apdx:sps-laaj-prompt}

The following are the different prompts used for the SPS Dataset:

``You will be given a context, a question, and an answer.
Evaluate the answer and score it on a rubrics of 6 criterias, including Conciseness, Interpretability, Completeness, Clarity, Accuracy, and Relevance, on a scale of -2 to 2. 
Just output 6 numbers, and do not provide any other explanation.

CONTEXT: {}

QUESTION: {}

ANSWER: {}

SCORES:
The 6 scores are: ''

\subsection{Training Hyperparameters} \label{apdx:hyperparams}

\begin{table}[ht]
\centering
\small
\begin{tabular}{lc}
\toprule
\textbf{Hyperparameter} & \textbf{Value} \\
\midrule
Quantization & 4-bit \\
LoRA Rank ($r$) & 16 \\
Precision & bfloat16 \\
Epochs & 5 \\
Batch Size & 32 \\
Gradient Accumulation & 1 \\
Learning Rate & $5 \times 10^{-5}$ \\
LR Scheduler & Linear \\
Warmup Ratio & 0.05 \\
Weight Decay & 0.01 \\
Hardware & NVIDIA A100 \\
\bottomrule
\end{tabular}
\caption{Hyperparameter settings for model fine-tuning.}
\label{tab:hyperparameters}
\end{table}

\subsection{Train/Test Loss Plot with and without Augmentation} \label{subsec: train/test loss plot augmentation}

\begin{figure}[!htbp]
    \centering
    \includegraphics[width=0.9\linewidth]{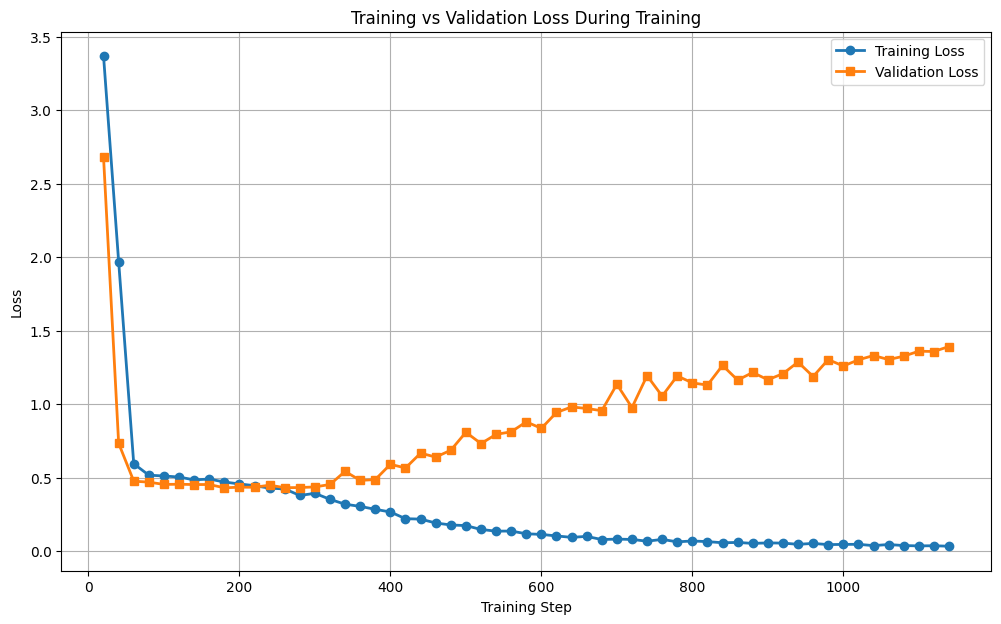}
    \caption{Plot of Training and Validation loss without using Data Augmentation}
\end{figure}

\begin{figure}[!htbp]
    \centering
    \includegraphics[width=0.9\linewidth]{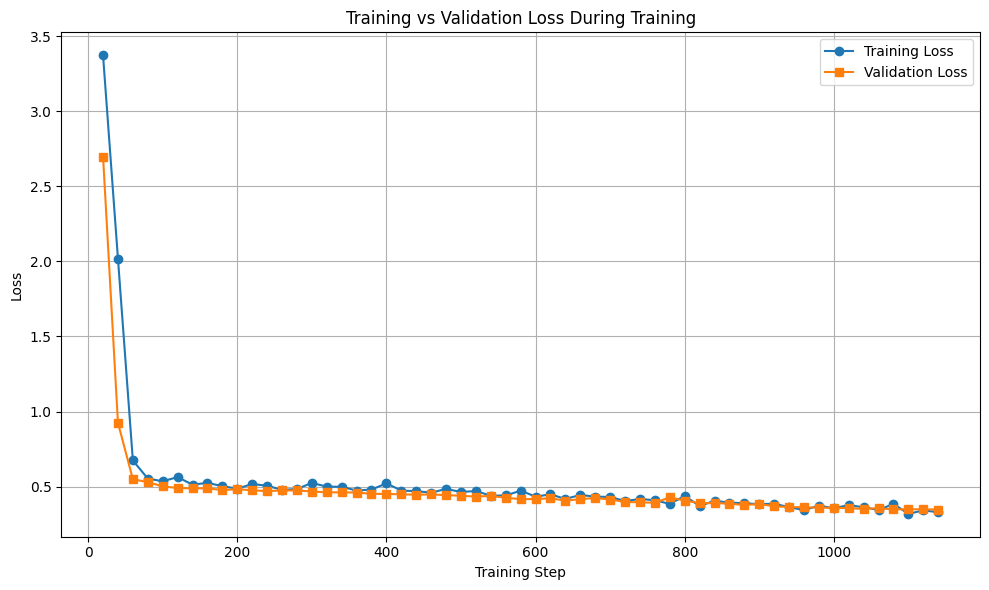}
    \caption{Plot of Training and Validation loss using Proposed Data Augmentation}
\end{figure}

\subsection{Train/Test Loss Plot with LoRA Dropout} \label{subsec: train/test loss plot lora dropout}

\begin{figure}[!htbp]
    \centering
    \includegraphics[width=0.9\linewidth]{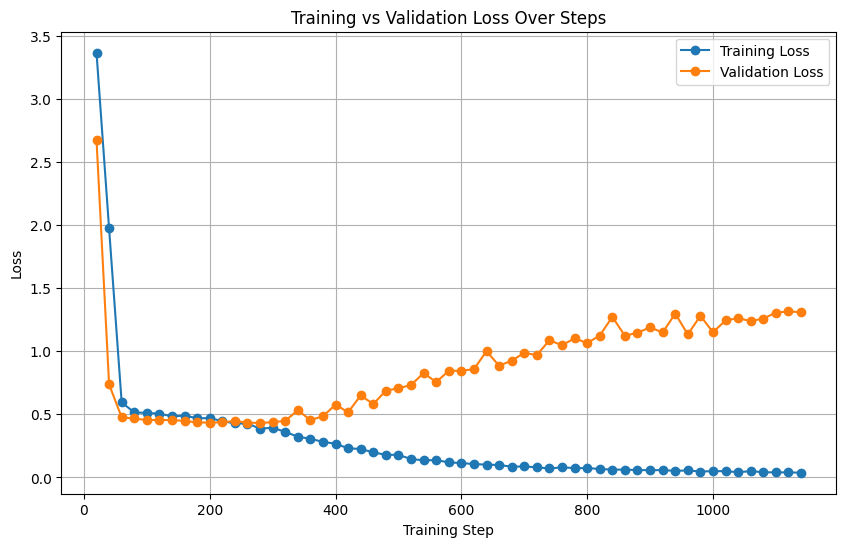}
    \caption{Plot of Training and Validation loss using LoRA Dropout}
\end{figure}

\end{document}